\documentclass[11pt,a4paper]{article}
\usepackage[hyperref]{acl2017}
\usepackage{times}
\usepackage{amsmath}
\usepackage{amssymb}
\usepackage{capt-of}
\usepackage{graphicx}
\usepackage{latexsym}
\usepackage{float}
\usepackage{tikz}
\usetikzlibrary{positioning}
\usetikzlibrary{bayesnet}
\usepackage[algoruled,linesnumbered,noend]{algorithm2e}
\usepackage[skip=5pt]{caption}
\graphicspath{ {./img/} }

\usepackage{url}

\aclfinalcopy 
\def\aclpaperid{14} 

\title{Grounding Symbols in Multi-Modal Instructions}

\author{Yordan Hristov, Svetlin Penkov, Alex Lascarides and Subramanian Ramamoorthy\\
  School of Informatics \\
  The University of Edinburgh \\
  {\tt\{yordan.hristov@, sv.penkov@, alex@inf., s.ramamoorthy@\}ed.ac.uk}}

\date{}

\begin{document}
\maketitle
\begin{abstract}
As robots begin to cohabit with humans in semi-structured environments, the need arises to understand instructions involving rich variability---for instance, learning to ground symbols in the physical world. Realistically, this task must cope with small datasets consisting of a particular users' contextual assignment of meaning to terms. We present a method for processing a raw stream of cross-modal input---i.e., linguistic instructions, visual perception of a scene and a concurrent trace of 3D eye tracking fixations---to produce the segmentation of objects with a correspondent association to high-level concepts. To test our framework we present experiments in a table-top object manipulation scenario. Our results show our model learns the user's notion of colour and shape from a small number of physical demonstrations, generalising to identifying physical referents for novel combinations of the words.
\end{abstract}



\section{Introduction}

\begin{figure}[t]
\centering
\includegraphics[scale=0.6]{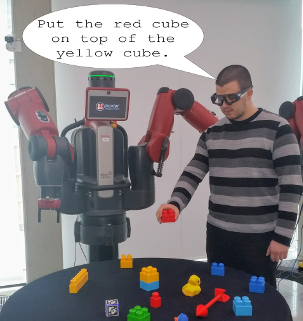}
\caption{Combining natural language input with eye tracking data allows for the dynamic labelling of images from the environment with symbols. The images are used in order to learn the meaning of language constituents and then ground them in the physical world.}
\label{fig:cover}
\end{figure}

Effective and efficient human-robot collaboration requires robots to interpret ambiguous instructions and concepts within a particular context, communicated to them in a manner that feels natural and unobtrusive to the human participant in the interaction. Specifically, the robot must be able to:
\begin{itemize}
\setlength\itemsep{0em}
\item Understand natural language instructions, which might be ambiguous in form and meaning.
\item Ground symbols occurring in these instructions within the surrounding physical world.
\item Conceptually differentiate between instances of those symbolic terms, based on features pertaining to their grounded instantiation, e.g. shapes and colours of the objects.
\end{itemize}

Being able to relate abstract symbols to observations with physical properties in the real world is known as the physical symbol grounding problem \cite{Vogt2002}; which is recognised as being one of the main challenges for human-robot interaction and constitutes the focus of this paper.

There is increasing recognition that the meaning of natural language words derives from how they manifest themselves across multiple modalities. Researchers have actively studied this problem from a multitude of perspectives. This includes works that explore the ability of agents to interpret natural language instructions with respect to a previously annotated semantic map \cite{matuszek2013mapping} or fuse high-level natural language inputs with low-level sensory observations in order to produce a semantic map \cite{wheelchairmap}.  \citet{matuszek2014gesture, baxterjesture} and \citet{kollar2013toward} tackle learning symbol grounding in language commands combined with gesture input in a table-top scenario. However, all these approaches depend on having {\textit{predefined}} specifications of different concepts in the environment: they either assume a pre-annotated semantic map with respect to which they ground the linguistic input or have an offline trained symbol classifier that decides whether a detected object can be labelled with a specific symbol; e.g. colour and shape in \cite{matuszek2014gesture}. Thus in order to deploy such a system, one should have access to an already trained classifier for every anticipated symbol, prior to any user interaction.
\begin{figure*}[t]
\centering
\includegraphics[scale=0.325]{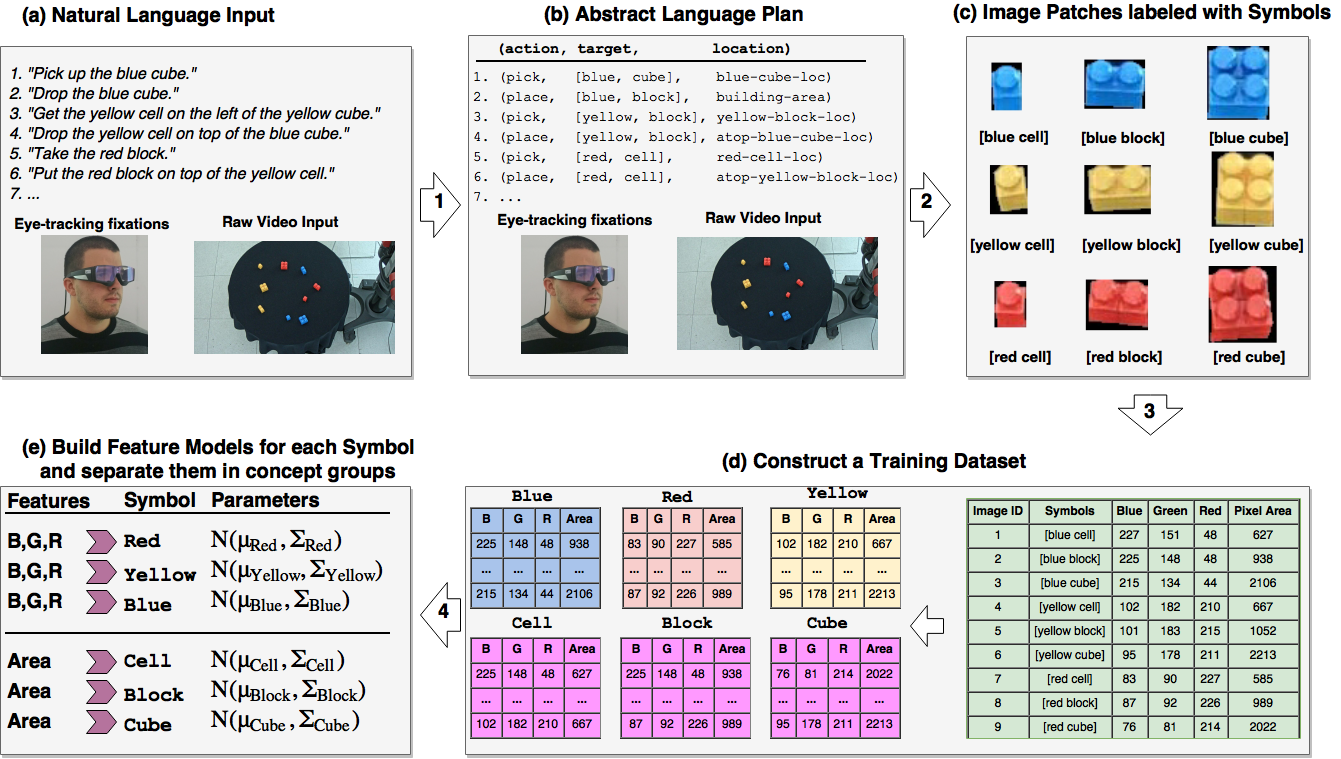}
\caption{Overview of the full system pipeline. Input to the system are natural language instructions, together with eye-tracking fixations and a camera view of the world from above \textbf{(a)} Natural language instructions are deterministically parsed to an abstract plan language \textbf{(b)} Using the abstract plan, a set of labelled image patches is produced from the eye-tracking and video data \textbf{(c)} Observable predefined features are extracted from the image patches \textbf{(d)} Each symbol is grounded to a subset of observable features \textbf{(e)}}
\label{fig:pipeline}
\end{figure*}

Multi-modal learning algorithms based on deep neural networks are also popular for grounding natural language instructions to the shared physical environment \cite{deepbolzman, multimodalspeech}.  But the majority of these algorithms depend crucially on large and pre-labelled datasets, and the challenge is in collecting these large-scale labelled datasets so that they not only capture the variability in language but also manage to represent the nuances (especially across multiple high-bandwidth modalities, such as vision and eye-tracking) of inter-personal variability in assignment of meaning (e.g., what one person calls {\em mustard} another might call {\em yellow}), which we claim is a key attribute of free-form linguistic instructions in human-robot interaction applications. If a previously unseen instruction/visual observation is presented to these systems, they might fail to ground or recognize them in the way that the user might have intended in that specific setting. \citet{spamfetch} potentially bypasses the need to collect a big dataset by demonstrating that a model trained in simulation can be successfully deployed on a robot in the real world. However, the problem is then shifted to generating task-specific training data in a simulator which approximates the real world well enough.

A proposed alternative to this off-line learning approach is to interactively teach an embodied agent about its surrounding world, assuming limited prior knowledge. \citet{al2016natural} demonstrates a model for incrementally learning the visual representation of words, but relies on temporally aligned videos with corresponding annotated natural language inputs. \citet{parde2015grounding} and \citet{thomason2016learning} represent the online concept learning problem as a variation of the interactive ``I Spy" game. However, these approaches assume an initial learning/exploratory phase in the world and extracted features are used as training data for all concept models associated with an object.

\citet{glide} introduce a method called GLIDE (see \S\ref{sec:glide} for details), which successfully teaches agents how to map abstract instructions, represented as a LISP-like program, into their physical equivalents in the world. Our work builds on this method: it uses it to achieve {\em natural language} symbol grounding, as a by-product of user interaction in a task-oriented scenario. Our approach achieves the following:
\begin{itemize}
\setlength\itemsep{0em}
\item It maps natural language instructions to a planned behaviour, such as in a robotic manipulation domain; in so doing it supports a communication medium that human users find natural. 
\item It learns symbol grounding by exploiting the concept of \textit{intersective modification} \cite{modificationbook} ---i.e., an object can be labelled with more than one symbol. The meaning of the symbols is learned with respect to the observed features of the instances of the object. 
\end{itemize}

In our work the agent assumes some prior knowledge about the world in the form of low-level features that it can extract from objects in the visual input---e.g. intensities in the primary colour channels and areas of pixel patches of any specific colour. On top of this, we learn classifiers for performing symbol grounding.  Each symbol has a probabilistic model which is fit to a subset of the extracted (visual) features. When a new instruction is received, the classifier for each symbol makes a decision regarding the object in the world (and their respective features) to which the symbol may be grounded. Crucially, the data from which these classifiers are learned is collected from demonstrations at `run time' and not prior to the specific human-robot interaction. Images of objects are extracted from the high-frequency eye tracking and video streams, while symbols that refer to these objects in the images are extracted from the parsed natural language instructions---see Figure \ref{fig:cover}. Through cross-modal instructions, the human participant is simultaneously teaching the robot how to execute a task and what properties the surrounding objects must have for that execution to be successful. For instance, while observing how to make a fruit salad in a kitchen, apart from learning the sequence of steps, the system would also gain an initial approximation of the visual appearance of different pieces of fruit and their associated natural language symbols.

\section{Methods}

Figure \ref{fig:pipeline} depicts the architecture of the overall system.  It consists of an end-to-end process, from raw linguistic and video inputs on the one hand to learned meanings of symbols that in turn are conceptually grouped: i.e., a symbol can correspond either to an object in the real world, or to a property of an object. The rest of the section is organized in the following fashion - each subsection corresponds to a numbered transition (1 to 4) indicated in Figure \ref{fig:pipeline}.

\subsection{Natural Language Semantic Parsing}

The task of the semantic parser is to map natural language requests into instructions represented in an abstract form. The abstract form we use is a list of tuples with the format \verb!(action target location)! (Figure \ref{fig:pipeline}b), where \verb!action! corresponds to an element from a predefined set $\mathcal{A}$, \verb!target! corresponds to a list of terms that describe an object in the world and \verb!location! corresponds to a single symbol denoting a physical location in the environment.

The narration of the plan execution by the human comprises one sentence per abstract instruction. Therefore, given a plan description, our semantic parser finds a mapping from each sentence to a corresponding instruction as defined by our abstract plan language.

\begin{figure*}[t]
\centering
\includegraphics[scale=0.7]{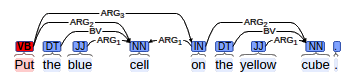}
\includegraphics[scale=0.7]{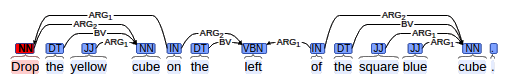}
\caption{Example of dependency graphs for input sentences. Red labels denote the top graph nodes.}
\label{fig:eds_compiled}
\end{figure*}

Elementary Dependency Structures (EDS) \cite{eds}, which are output by parsing the sentence with the wide-coverage English Resource Grammar \cite{erg}, are used as an intermediate step in this mapping procedure. EDS are given as dependency graphs (Figure \ref{fig:eds_compiled}) and are a variable-free reduced form of the full Minimal Recursion Semantics (MRS) \cite{mrs} representation of the natural language input.  Given EDS for a particular sentence, parsing proceeds in two steps:
\begin{itemize}
\setlength\itemsep{0em}
\item The graph is extracted from nodes and their respective edges, cleaning up nodes and edges that do not contribute directly to the meaning of the instruction---i.e., anything that is not a verb, adjective, preposition or a noun;
\item The processed graph is recursively traversed until all nodes have been visited at least once.
\end{itemize}

Knowing the format of our abstract plan language, we know that \verb!action! would always correspond to the verb in the input sentence, \verb!target! would correspond to a noun phrase and \verb!location! would correspond to a prepositional phrase (i.e., a combination of a preposition and noun phrase). For us, the noun phrases all consist of a noun, complemented by a possibly empty list of adjectives. Extracting the action is straightforward since the top node in the EDS always corresponds to the verb in the sentence; see Figure \ref{fig:eds_compiled}. The extracted action is then passed through a predefined rule-based filter which assigns it one of the values from $\mathcal{A}$: e.g. \textit{pick, grab, take, get} would all be interpreted as \verb!pick!.

The \verb!target! entry can be extracted by identifying noun node in the EDS that's connected to the verb node. Once such a noun is found, one can identify its connections to any adjective nodes---this gives a full list of symbols that define the object referenced by the \verb!target!. 

The \verb!location! entry can be extracted by searching for preposition nodes in the EDS that are connected to the verb node. If there is no such node, then the \verb!location! is constructed directly from the target by concatenating its labels - e.g. for a blue cube the \verb!location! would be the symbol \verb!blue-cube-location!. Extracting the \verb!location! from a prepositional phrase is less constrained since different verbs can be related to spatial prepositions in varied ways---either the preposition node has an edge connecting it to the verb node or vice versa. Once a prepositional node is visited, we proceed by recursively exploring any chains of interconnected nouns, prepositions, and adjectives. The recursion calls for backtracking whenever a node is reached with no unvisited incoming or outgoing edges: e.g., node \verb!cube! on Figure \ref{fig:eds_compiled} (bottom). For example, the symbol \verb!on-left-of-cube! is produced for \verb!location! for the bottom sentence in Figure \ref{fig:eds_compiled}.

In this way, the result of parsing is a sequence of abstract instructions---i.e., an abstract plan---together with a {\em symbol set} $S$, containing all symbols which are part of any \verb!target! entry. At this point, the symbols are still not actually grounded in the real world. Together with the raw video feed and the eye-tracking fixations, the abstract plan becomes an input to GLIDE \cite{glide}.

\subsection{Grounding and Learning Instances through Demonstration and Eye tracking (GLIDE)}
\label{sec:glide}

\citet{glide} introduce a framework for Grounding and Learning Instances through Demonstration and Eye tracking (GLIDE). In this framework, {\textit{fixation programs}}  are represented in terms of fixation traces obtained during task demonstrations combined with a high-level plan. Through probabilistic inference over fixation programs, it becomes possible to infer latent properties of the environment and determine locations in the environment which correspond to each instruction in an input abstract plan that conforms to the format discussed above - \verb!(action target location)!.

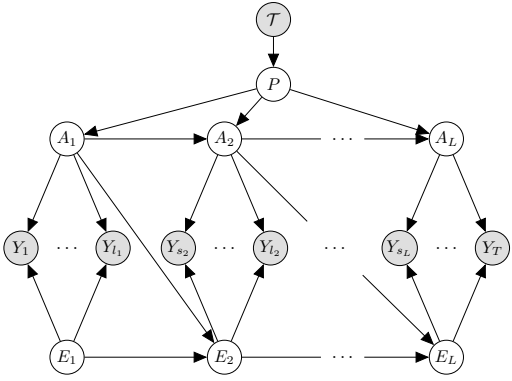
\begin{figure}
	\centering
	\scalebox{.8}{\begin{tikzpicture}
  [scale=0.8, every node/.style={scale=0.8}]

  \node[obs] (Y11) {$Y_{1}$};
  \node[const, right=0.2cm of Y11] (Y1l) {$\;\ldots\;$};
  \node[obs, right=0.2cm of Y1l] (Y1L)  {$Y_{l_1}$};
  
  \node[obs, right=0.5cm of Y1L] (Y21) {$Y_{s_2}$};
  \node[const, right=0.2cm of Y21] (Y2l) {$\;\ldots\;$};
  \node[obs, right=0.2cm of Y2l] (Y2L)  {$Y_{l_2}$};
  
  \node[const, right=0.5cm of Y2L] (Yt) {$\;\ldots\;$};
  
  \node[obs, right=0.5cm of Yt] (Y31) {$Y_{s_L}$};
  \node[const, right=0.2cm of Y31] (Y3l) {$\;\ldots\;$};
  \node[obs, right=0.2cm of Y3l] (Y3L)  {$Y_{T}$};
  
  \node[latent, above=1.5cm of Y1l]		(A1)      {$A_1$};
  \node[latent, above=1.5cm of Y2l]		(A2)      {$A_2$};
  \node[latent, above=1.5cm of Y3l]		(AL)      {$A_L$};
  
  \node[latent, below=1.5cm of Y1l]		(E1)      {$E_1$};
  \node[latent, below=1.5cm of Y2l]		(E2)      {$E_2$};
  \node[latent, below=1.5cm of Y3l]		(EL)      {$E_L$};
  
  \edge[]{A1}{A2}
  \edge[]{A1}{E2}
  \edge[]{A1}{Y11}
  \edge[]{A1}{Y1L}

  \edge[]{E1}{E2}
  \edge[]{E1}{Y11}
  \edge[]{E1}{Y1L}
  
  \node[const, right=1.3cm of A2]  (INF1) {$\;\;\ldots\;\;$};
  \edge[-]{A2}{INF1}
  \node[const, above right=0.2cm and 0.4cm of Y2L]  (INF2) {};
  \edge[-]{A2}{INF2}
  \edge[]{A2}{Y21}
  \edge[]{A2}{Y2L}

  \node[const, right=1.3cm of E2]  (INF3) {$\;\;\ldots\;\;$};
  \edge[-]{E2}{INF3}
  \edge[]{E2}{Y21}
  \edge[]{E2}{Y2L}
  
  \edge[]{INF1}{AL}
  \node[const, below left=0.2cm and 0.4cm of Y31]  (INF4) {};
  \edge[]{INF4}{EL}
  \edge[]{AL}{Y31}
  \edge[]{AL}{Y3L}

  \edge[]{INF3}{EL}
  \edge[]{EL}{Y31}
  \edge[]{EL}{Y3L}
  
  \node[latent, above right=0.5cm and 0.4cm of A2] (P) {$P$};
  \edge[]{P}{A1}
  \edge[]{P}{A2}
  \edge[]{P}{AL}
  
  \node[obs, above=0.5cm of P] (T) {$\mathcal{T}$};
  \edge[]{T}{P}

\end{tikzpicture}}
    \caption{The proposed probabilistic model for physical symbol grounding is based on the idea that ``task and context determine where you look'' \cite{Rothkopf2007}.}
    \label{fig:pgm}
\end{figure}

\subsubsection{3D Eye-Tracking}

Mobile eye trackers provide fixation information in pixel coordinates corresponding to locations in the image of a first person view camera. In order to utilise information from multiple input sensors, an additional camera may be attached to augment an eye-tracker, by running mono camera SLAM algorithm in the background---ORBSLAM \cite{Mur2015}. The SLAM algorithm provides 6D pose of the eye tracking glasses within the world frame; this allows for the fixation locations to be projected into the 3D world by ray casting and finding intersections with a 3D model of the environment. As a result, fixations can be represented as 3D locations, enabling the projection of fixations in the frame of any sensor in the environment. 

\subsubsection{Model and Location Inference}

In order to solve the problem of symbol grounding, inference is performed using a generative probabilistic model, which is shown in Figure \ref{fig:pgm}. The sequence of fixations $Y_1:Y_T$ depend both on the current environment state $E$ and the action being executed $A$. Each action is part of the plan $P$ which is determined by the task being demonstrated $T$. The sequence of fixations is observed and interpreted with respect to a task that is by this stage already known, while the state of the environment and the current action are unknown. The main inference task is to determine the structure of the model and assign each fixation to the action that is its cause. A crucial feature of this procedure is the fact, deriving from human sensorimotor behaviour, that the distribution of fixations is different when a person is attending to the execution of an action compared to periods of transitions between actions in the plan. By utilising this property and using samples from a Dirichlet distribution to describe these transition points, GLIDE is able to infer the correct partitioning of the fixation sequence. This information allows us to localise each item of interest in the environment and extract labelled sensory signals from the environment. A complete description of this model and inference procedure can be found in \cite{glide}.

\subsection{Feature Extraction}

The parser produces a set of symbols $S$ and GLIDE produces a set of image patches $I$, each of which is labelled with a subset of symbols from $S$. We proceed by extracting a number of features, drawn from a pre-existing set $F$. The features are derived from the objects in the image patches, after removing the background. This is achieved through a standard background subtraction method---we know that the majority of the image will be occupied by a solid object with a uniform colour, anything else is a background.  For instance, in the image patches in Figure \ref{fig:pipeline} (c), the objects are the colourful blocks and the background is the black strips around them. Images containing only or predominantly background are considered noise in the dataset and are discarded. For each symbol $s$ we group the extracted features from each image labelled with $s$ resulting in $S$ lists of $M_s$ tuples with $F$ entries in each tuple, where $M_s$ is the number of images being labelled with $s$; see Figure \ref{fig:pipeline} (d, left). The data for each feature is normalized to fall between 0 and 1.

\begin{algorithm}[t]
\caption{Symbol Meaning Learning}
	\KwIn{$\sigma_{thresh}$}
	\KwData{$I$, $S$, $F$}
    \KwOut{$K = \{(\mu_1, \Sigma_1), \ldots ,(\mu_S, \Sigma_S)\}$, $C = \{ (F_{invar}^{s} : s), \ldots (F_{invar}^{S} : S) \}$} \BlankLine
	$Data \leftarrow [s_1:\{\}, \ldots, s_S:\{\}]$\;
  	\For{image $i$ in $I$}{
  		$symbols_{i}  \leftarrow GetSymbols(i)$\;
    	$features_{i}  \leftarrow ExtractFeatures(i)$\;
    	\For{$symbol$ in $symbols_{i}$}{
    		Append $features_{i}$ to $Data[symbol]$\;
    	}
  	}
    \For{$s$ in $S$}{
    		$K_{s} \leftarrow FitNormal(Data[s])$\;
            $K_{s} \leftarrow CleanNoise(Data[s])$\;
            $F_{invar}^{s} \leftarrow FindInvFeat(K_{s}, \sigma_{thresh})$\;
            $K_{s} \leftarrow RefitNormal(K_{s}, F_{invar}^{s})$\;
            Append $(F_{invar}^{s} : K_{s})$ to $C$\;
    }
  	\label{alg:meaning_learning}
\end{algorithm}

\subsection{Symbol Meaning Learning}

\begin{figure}[t]
\centering
\includegraphics[scale=0.3]{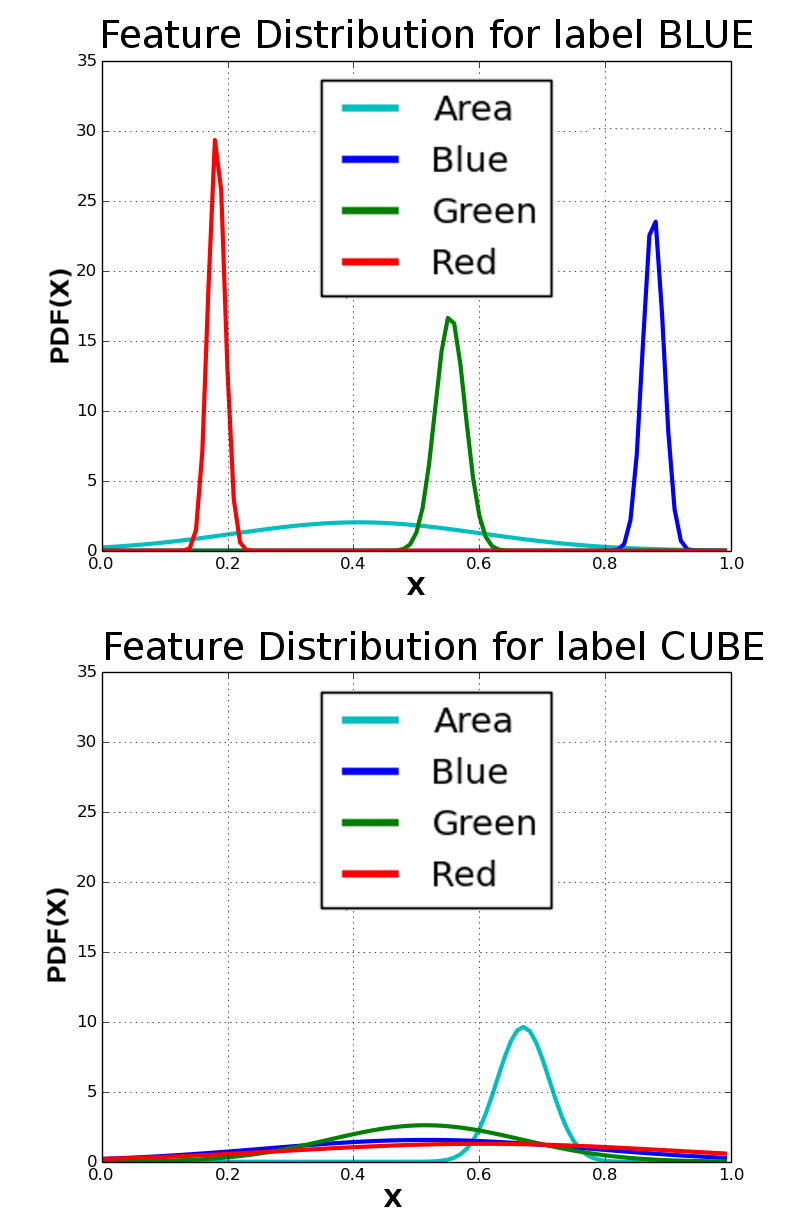}
\caption{Example of feature distributions for \texttt{blue} (top) and \texttt{cube} (bottom) symbols.}
\label{fig:distributions}
\end{figure}

For each symbol $s \in S$ and each feature $f \in F$ we fit a 1-D Normal distribution resulting in a new list of tuples with size $F$ - $s_j:[(\mu_{f_1}^{s_j}, \sigma_{f_1}^{s_j}),\ldots,(\mu_{f_F}^{s_j}, \sigma_{f_F}^{s_j})]$ for the $j^{th}$ symbol. Taking into account that the object location process in GLIDE could still produce noisy results---i.e., the label of an image can be associated with the wrong symbol---we process our distributions to refit them to data that falls within two standard deviations from the means of the original distributions. We are then seeking observed features $f$ that are invariant with respect to each token use of a specific symbol $s$ within the user instructions so far---i.e. their distributions are `narrow' and with variance below a predefined threshold $\sigma_{thresh}$ (see Figure \ref{fig:distributions}). If we have a set of images that are of blue objects with different shapes and we extract a set of features from them, we would expect that features with lower variation (e.g. RGB channels as opposed to area) would explain the colour blue better than features with more variation (i.e. pixel area).

In the last step, we construct a set of the invariant features from the discovered narrow distributions for a given symbol $l$ - $(F_{invar}^{s})$ - and say that this set characterizes the symbol. The parameters for the symbol are the concatenation of the means of the features from $(F_{invar}^{l})$ into a mean vector and the concatenation of the variances into a diagonal covariance matrix. The resultant mean vector and covariance matrix are later used for inference when shown a new set of images.

\section{Experiments}

We now present results from initial experiments based on the framework in Figure \ref{fig:pipeline}. We focus our explanation on steps 3 and 4 in that figure, as these are the pertinent and novel elements introduced here. The input data for Figure \ref{fig:pipeline} (c) is derived from the process already well described in \cite{glide}.

\subsection{Dataset}

For our experiments we used a total of six symbols defining $S$: 3 for colour (red, blue and yellow); and 3 for shape (cell, block, cube).  We used four extracted features for $F$: R, G, B values and pixel area. The objects used were construction blocks that can be stacked together and images of them were gathered in a tabletop robotic manipulation setup (see Figure \ref{fig:pipeline} (a)). Based on the empirical statistics of the recognition process in \cite{glide}, our input dataset to the Symbol Meaning Learning algorithm consists of 75\% correctly annotated and 25\% mislabelled images. The total training dataset comprised of approximately 2000 labelled image patches, each of which is labelled with two symbols---e.g. \verb!blue cell!, \verb!red block!, \verb!yellow cube!, etc.

\begin{figure}[t]
\centering
\includegraphics[scale=0.12]{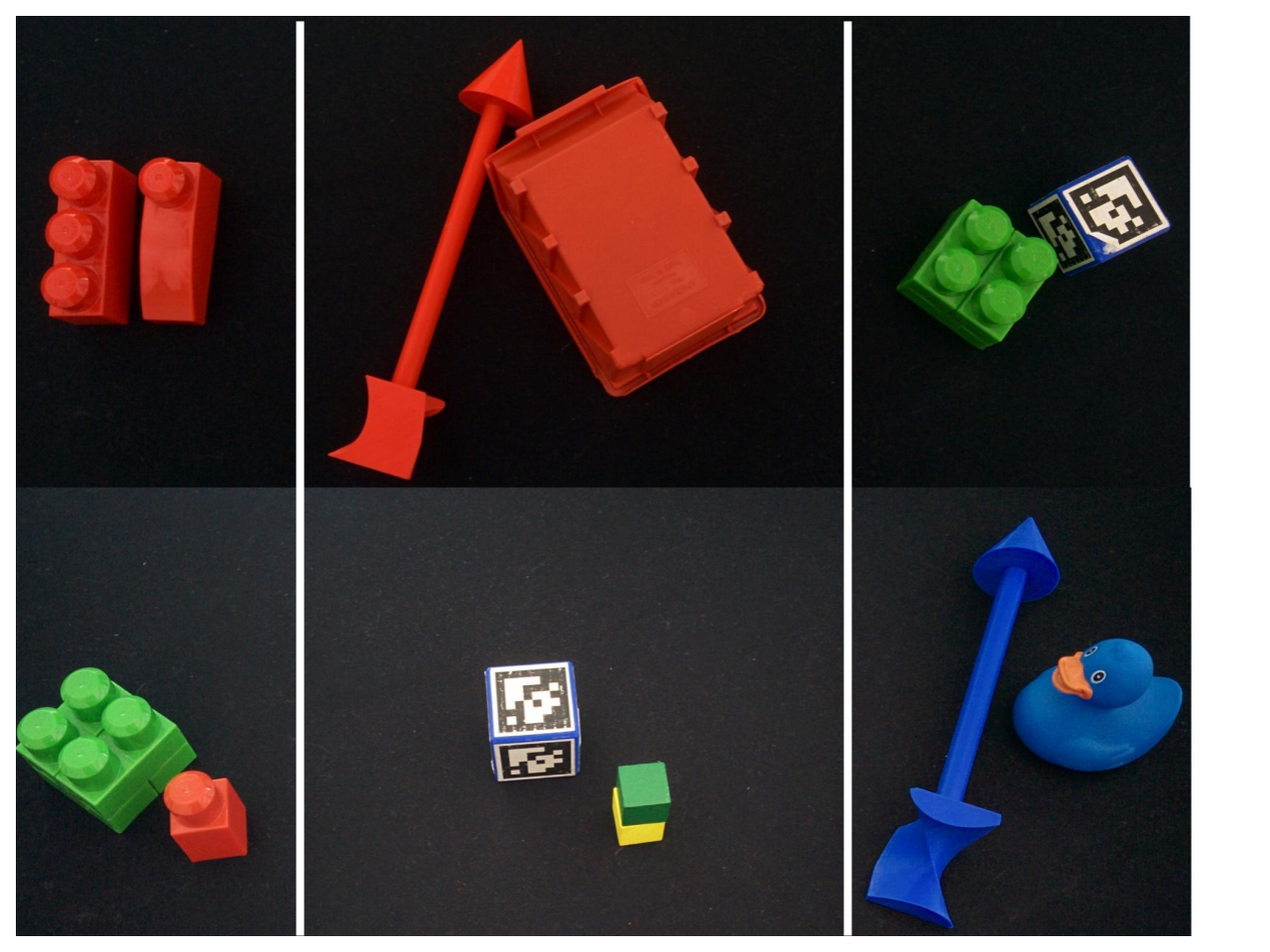}
\caption{Variations in the objects from the test set for colour (top half) and shape (bottom half)}
\label{fig:new_objects}
\end{figure}

The additional test set was designed in two parts: one that would test colour recognition and one that would test shape recognition. Overall, 48 objects were presented to the algorithm where the features for each object would fall into one of the following categories:

\begin{itemize}
\setlength\itemsep{0em}
\item Previously seen features (Figure \ref{fig:new_objects} (left))
\item Previously unseen features, close to the features of the training data (Figure \ref{fig:new_objects} (middle))
\item Previously unseen features, not close to the features of the training data (Figure \ref{fig:new_objects} (right))
\end{itemize}

\subsection{Experimental Set up}

Inference over new images is performed by thresholding the probability density function (PDF) values from the model parameters for each symbol. The idea is to test how well the algorithm can differentiate the learned concepts with slight variations from concepts it has not seen before: e.g. given that the algorithm was trained on 3 colours and 3 shapes, we would expect that it should recognize different hues of the 3 colours and objects with similar shapes to the original 3; however, it may not be able to recognize objects with completely different features. Moreover, we further group different symbols into concept groups. If any two symbols are described by the same features, it is safe to assume that those two symbols are mutually exclusive: that is, they can not both describe an object simultaneously. Thus we go over each concept group and if there are symbols yielding PDF values above a predefined threshold, we assign the new image the symbol from that group with the highest PDF.

\subsection{Results}

\begin{figure*}[t]
\centering
\includegraphics[scale=0.4]{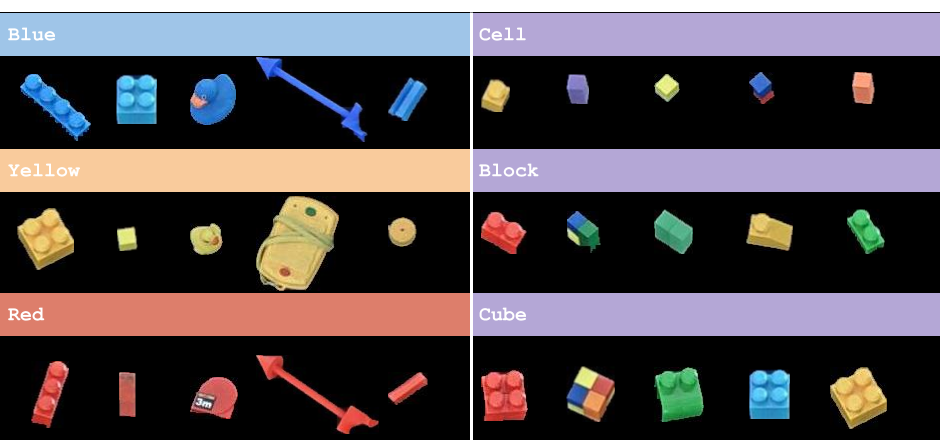}
\caption{Excerpt from the testing dataset of objects whose colour and shape were correctly recognised.}
\label{fig:results}
\end{figure*}

\begin{figure}[t]
\centering
\includegraphics[scale=0.3]{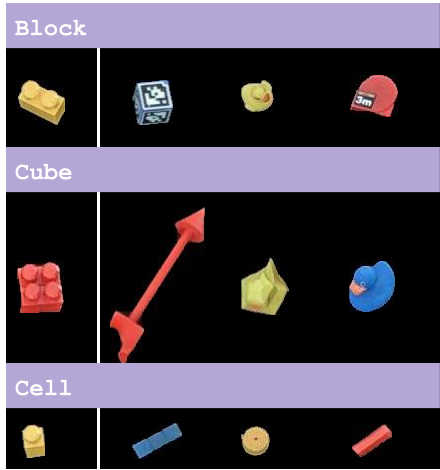}
\caption{Examples of wrongly assigned known symbols to unseen shapes. Leftmost objects demonstrate an object from the training data.}
\label{fig:wrong}
\end{figure}

The system successfully learns from the training dataset that the colour symbols are being characterized by the extracted RGB values, while (in contrast) the shape symbols from the pixel area of the image patch---see Figure \ref{fig:results}. Given a new test image with its extracted features, the algorithm recognises 93\% of presented colours and 56\% of presented shapes. Tables \ref{tab:cmcolor} and \ref{tab:cmshape} report the confusion matrices for the testing set.  This shows that the system is more robust when recognizing colours than when recognizing shapes. This can be attributed to the fact that while RGB values describe the concept of colour well enough, simply the pixel area is not enough to describe the concept of shape. Therefore the algorithm confuses the rubber duck with a cell, for example, and the arrow with a cube, see Figure \ref{fig:wrong}, principally because they are of a similar size to each other! In future work, we would consider a wider range of features being extracted from the images, which in turn would support a finer-grained discrimination among objects. 

\begin{center}
\begin{table}
\captionof{table}{Confusion matrix for colour symbols} \label{tab:cmcolor}
    \begin{tabular}{| c | c | c | c | c |}
    \hline
      & \textbf{Red} & \textbf{Yellow} & \textbf{Blue} & \textbf{Unknown} \\ \hline
    \textbf{Red} & \textbf{12} & 1 & 0 & 0 \\ \hline
    \textbf{Yellow}  & 0 & \textbf{12} & 0 & 0 \\ \hline
    \textbf{Blue} & 0 & 0 & \textbf{13} & 1 \\ \hline
    \textbf{Unknown} & 0 & 1 & 0 & \textbf{13} \\ \hline
    \end{tabular}

\captionof{table}{Confusion matrix for shape symbols} \label{tab:cmshape}
    \begin{tabular}{| c | c | c | c | c |}
    \hline
      & \textbf{Cell} & \textbf{Block} & \textbf{Cube} & \textbf{Unknown} \\ \hline
    \textbf{Cell} & \textbf{8} & 1 & 0 & 0 \\ \hline
    \textbf{Block}  & 1 & \textbf{7} & 0 & 0 \\ \hline
    \textbf{Cube} & 0 & 3 & \textbf{8} & 0 \\ \hline
    \textbf{Unknown} & 5 & 5 & 6 & \textbf{4} \\ \hline
    \end{tabular}
\end{table}
\end{center}

\section{Discussion and Future Work}
The experiments in this paper have demonstrated that it is possible to train classifiers for object appearance alongside symbols, which are analysed via a semantic parser, to achieve grounding of instructions that respect the specificity of the scenario within which that association derives its meaning. Although our framework supports an entire pipeline, from raw cross-modal input to an interpreted and grounded instruction, the presented scenarios are simple and the specific methods could be made (much) more sophisticated. Despite this, we claim that this provides stepping stones towards learning more complex language structures in the future: during the first few demonstrations a human could teach a robot fundamental concepts like colours, shapes, orientation, and then proceed to use this newly gained knowledge to ground, e.g., prepositional phrases  \cite{forbes2015robot,rosman2011learning} or actions \cite{tellmedave,zampogiannis2015learning}  in an online and context specific manner. Once the system knows what blue cubes look like, it would be easier to learn what it means for another cube to be on top of/around it.

Another fruitful line of exploration would be continuous learning of both known and unknown symbols, using the same conceptual groups the system has extracted from the training data.  For instance, whenever the robot observes a new object it can either label it with a symbol or deem it as unknown for a particular concept group. Whenever a symbol is assigned, the feature model for that symbol is updated, taking into account the new data point. If on the other hand the symbol is unknown, the system can prompt the human for new linguistic input which together with its feature model is added to the knowledge base and allows for its future recognition. For example, if the robot observes a new hue of blue it would update its parameters for \verb!blue! to account for that; whereas if it observes a new colour (e.g. green) it would ask the human for the unknown symbol and would record it for future reference.

The idea of teaching the system about compound nouns is also a relevant challenge and a possible extension of this work: our current setup relies on noun phrases consisting of predicative \verb!Adj!s and a \verb!Noun! (e.g. blue cube), and so we know that the associated image patch $X$ satisfies both the adjective and the noun---i.e., \verb!blue!(X) and \verb!cube!(X) are both true. However, this would not apply to a compound noun like steak knife: we know that the associated image patch $X$ satisfies \verb!knife!(X) but does not satisfy \verb!steak!(X). Refinements to our model would be necessary in order to represent more complex symbol relations, e.g. in a hierarchical fashion \cite{new_objects}.

\section{Conclusion}
We present a framework for using cross-modal input: a combination of natural language instructions, video and eye tracking streams, to simultaneously perform semantic parsing and grounding of symbols used in that process within the physical environment. This is achieved without reliance on pre-existing object models, which may not be particularly representative of the specifics of a particular user's contextual usage and assignment of meaning within that rich multi-modal stream. Instead, we present an online approach that exploits the pragmatics of human sensorimotor behaviour to derive cues that enable the grounding of symbols to objects in the stream. Our preliminary experiments demonstrate the usefulness of this framework, showing how a robot is not only able to learn a human's notion of colour and shape, but also that it is able to generalise to the recognition of these features in previously unseen objects from a small number of physical demonstrations.

\section*{Acknowledgments}

This work is partly supported by ERC Grant 269427 (STAC), a Xerox University Affairs Committee grant, and grants EP/F500385/1 and BB/F529254/1 for the DTC in Neuroinformatics and Computational Neuroscience from the UK EPSRC, BBSRC, and the MRC.  We are very grateful for the feedback from anonymous reviewers.


\bibliography{acl2017}

\begin{thebibliography}{}
\expandafter\ifx\csname natexlab\endcsname\relax\def\natexlab#1{#1}\fi

\bibitem[{Al-Omari et~al.(2016)Al-Omari, Duckworth, Hogg, and
  Cohn}]{al2016natural}
M~Al-Omari, P~Duckworth, DC~Hogg, and AG~Cohn. 2016.
\newblock Natural language acquisition and grounding for embodied robotic
  systems.
\newblock In {\em Proceedings of the 31st Association for the Advancement of
  Artificial Intelligence\/}. AAAI Press.

\bibitem[{Copestake et~al.(2005)Copestake, Flickinger, Pollard, and Sag}]{mrs}
Ann Copestake, Dan Flickinger, Carl Pollard, and Ivan~A Sag. 2005.
\newblock Minimal recursion semantics: An introduction.
\newblock {\em Research on Language and Computation\/} 3(2-3):281--332.

\bibitem[{Eldon et~al.(2016)Eldon, Whitney, and Tellex}]{baxterjesture}
Miles Eldon, David Whitney, and Stefanie Tellex. 2016.
\newblock Interpreting multimodal referring expressions in real time.
\newblock In {\em International Conference on Robotics and Automation\/}.

\bibitem[{Flickinger et~al.(2014)Flickinger, Bender, and Oepen}]{erg}
Dan Flickinger, Emily~M. Bender, and Stephan Oepen. 2014.
\newblock \href{http://www.delph-in.net/esd}{{ERG} semantic documentation}.
\newblock Accessed on 2017-04-29.
\newblock \href{http://www.delph-in.net/esd}{http://www.delph-in.net/esd}.

\bibitem[{Forbes et~al.(2015)Forbes, Rao, Zettlemoyer, and
  Cakmak}]{forbes2015robot}
Maxwell Forbes, Rajesh~PN Rao, Luke Zettlemoyer, and Maya Cakmak. 2015.
\newblock Robot programming by demonstration with situated spatial language
  understanding.
\newblock In {\em Robotics and Automation (ICRA), 2015 IEEE International
  Conference on\/}. IEEE, pages 2014--2020.

\bibitem[{Kollar et~al.(2013)Kollar, Krishnamurthy, and
  Strimel}]{kollar2013toward}
Thomas Kollar, Jayant Krishnamurthy, and Grant~P Strimel. 2013.
\newblock Toward interactive grounded language acqusition.
\newblock In {\em Robotics: Science and Systems\/}.

\bibitem[{Matuszek et~al.(2014)Matuszek, Bo, Zettlemoyer, and
  Fox}]{matuszek2014gesture}
Cynthia Matuszek, Liefeng Bo, Luke Zettlemoyer, and Dieter Fox. 2014.
\newblock Learning from unscripted deictic gesture and language for human-robot
  interactions.
\newblock In {\em AAAI\/}. pages 2556--2563.

\bibitem[{Matuszek et~al.(2013)Matuszek, Herbst, Zettlemoyer, and
  Fox}]{matuszek2013mapping}
Cynthia Matuszek, Evan Herbst, Luke Zettlemoyer, and Dieter Fox. 2013.
\newblock Learning to parse natural language commands to a robot control
  system.
\newblock In {\em Experimental Robotics\/}. Springer, pages 403--415.

\bibitem[{Misra et~al.(2016)Misra, Sung, Lee, and Saxena}]{tellmedave}
Dipendra~K Misra, Jaeyong Sung, Kevin Lee, and Ashutosh Saxena. 2016.
\newblock Tell me dave: Context-sensitive grounding of natural language to
  manipulation instructions.
\newblock {\em The International Journal of Robotics Research\/}
  35(1-3):281--300.

\bibitem[{Morzycki(2013)}]{modificationbook}
Marcin Morzycki. 2013.
\newblock \href{http://msu.edu/~morzycki/work/book}{Modification}.
\newblock Book manuscript. In preparation for the Cambridge University Press
  series \emph{Key Topics in Semantics and Pragmatics}.
\newblock
  \href{http://msu.edu/~morzycki/work/book}{http://msu.edu/~morzycki/work/book}.

\bibitem[{Mur-Artal et~al.(2015)Mur-Artal, Montiel, and Tard\'os}]{Mur2015}
Ra\'ul Mur-Artal, J.~M.~M. Montiel, and Juan~D. Tard\'os. 2015.
\newblock \href{https://doi.org/10.1109/TRO.2015.2463671}{{ORB-SLAM}: a
  versatile and accurate monocular {SLAM} system}.
\newblock {\em IEEE Transactions on Robotics\/} 31(5):1147--1163.
\newblock
  \href{https://doi.org/10.1109/TRO.2015.2463671}{https://doi.org/10.1109/TRO.2015.2463671}.

\bibitem[{Ngiam et~al.(2011)Ngiam, Khosla, Kim, Nam, Lee, and
  Ng}]{multimodalspeech}
Jiquan Ngiam, Aditya Khosla, Mingyu Kim, Juhan Nam, Honglak Lee, and Andrew~Y
  Ng. 2011.
\newblock Multimodal deep learning.
\newblock In {\em Proceedings of the 28th international conference on machine
  learning (ICML-11)\/}. pages 689--696.

\bibitem[{Oepen et~al.(2004)Oepen, Flickinger, Toutanova, and Manning}]{eds}
Stephan Oepen, Dan Flickinger, Kristina Toutanova, and Christopher~D Manning.
  2004.
\newblock {\em Research on Language and Computation\/} 2(4):575--596.

\bibitem[{Parde et~al.(2015)Parde, Hair, Papakostas, Tsiakas, Dagioglou,
  Karkaletsis, and Nielsen}]{parde2015grounding}
Natalie Parde, Adam Hair, Michalis Papakostas, Konstantinos Tsiakas, Maria
  Dagioglou, Vangelis Karkaletsis, and Rodney~D Nielsen. 2015.
\newblock Grounding the meaning of words through vision and interactive
  gameplay.
\newblock In {\em IJCAI\/}. pages 1895--1901.

\bibitem[{Penkov et~al.(2017)Penkov, Bordallo, and Ramamoorthy}]{glide}
Svetlin Penkov, Alejandro Bordallo, and Subramanian Ramamoorthy. 2017.
\newblock Physical symbol grounding and instance learning through demonstration
  and eye tracking .

\bibitem[{Rosman and Ramamoorthy(2011)}]{rosman2011learning}
Benjamin Rosman and Subramanian Ramamoorthy. 2011.
\newblock Learning spatial relationships between objects.
\newblock {\em The International Journal of Robotics Research\/}
  30(11):1328--1342.

\bibitem[{Rothkopf et~al.(2007)Rothkopf, Ballard, and Hayhoe}]{Rothkopf2007}
Constantin~A Rothkopf, Dana~H Ballard, and Mary~M Hayhoe. 2007.
\newblock Task and context determine where you look.
\newblock {\em Journal of vision\/} 7.

\bibitem[{Srivastava and Salakhutdinov(2012)}]{deepbolzman}
Nitish Srivastava and Ruslan~R Salakhutdinov. 2012.
\newblock Multimodal learning with deep boltzmann machines.
\newblock In {\em Advances in neural information processing systems\/}. pages
  2222--2230.

\bibitem[{Sun et~al.(2014)Sun, Bo, and Fox}]{new_objects}
Yuyin Sun, Liefeng Bo, and Dieter Fox. 2014.
\newblock Learning to identify new objects.
\newblock In {\em Robotics and Automation (ICRA), 2014 IEEE International
  Conference on\/}. IEEE, pages 3165--3172.

\bibitem[{Thomason et~al.(2016)Thomason, Sinapov, Svetlik, Stone, and
  Mooney}]{thomason2016learning}
Jesse Thomason, Jivko Sinapov, Maxwell Svetlik, Peter Stone, and Raymond~J
  Mooney. 2016.
\newblock Learning multi-modal grounded linguistic semantics by playing “i
  spy”.
\newblock In {\em Proceedings of the Twenty-Fifth international joint
  conference on Artificial Intelligence (IJCAI)\/}.

\bibitem[{Tobin et~al.(2017)Tobin, Fong, Ray, Schneider, Zaremba, and
  Abbeel}]{spamfetch}
Josh Tobin, Rachel Fong, Alex Ray, Jonas Schneider, Wojciech Zaremba, and
  Pieter Abbeel. 2017.
\newblock Domain randomization for transferring deep neural networks from
  simulation to the real world.
\newblock {\em arXiv preprint arXiv:1703.06907\/} .

\bibitem[{Vogt(2002)}]{Vogt2002}
Paul Vogt. 2002.
\newblock The physical symbol grounding problem.
\newblock {\em Cognitive Systems Research\/} 3(3):429--457.

\bibitem[{Walter et~al.(2014)Walter, Hemachandra, Homberg, Tellex, and
  Teller}]{wheelchairmap}
Matthew~R Walter, Sachithra Hemachandra, Bianca Homberg, Stefanie Tellex, and
  Seth Teller. 2014.
\newblock A framework for learning semantic maps from grounded natural language
  descriptions.
\newblock {\em The International Journal of Robotics Research\/}
  33(9):1167--1190.

\bibitem[{Zampogiannis et~al.(2015)Zampogiannis, Yang, Ferm{\"u}ller, and
  Aloimonos}]{zampogiannis2015learning}
Konstantinos Zampogiannis, Yezhou Yang, Cornelia Ferm{\"u}ller, and Yiannis
  Aloimonos. 2015.
\newblock Learning the spatial semantics of manipulation actions through
  preposition grounding.
\newblock In {\em Robotics and Automation (ICRA), 2015 IEEE International
  Conference on\/}. IEEE, pages 1389--1396.

\end{thebibliography}
\bibliographystyle{acl_natbib}

\end{document}